\documentclass[]{tGIS2e}
\usepackage{color}
\usepackage{underscore}
\usepackage{url}

\begin{document}

\title{Urban land-use analysis using proximate sensing imagery: a survey}

\author{
Zhinan Qiao and Xiaohui Yuan 
\affil{University of North Texas, 1155 Union Circle, Denton, TX 76203}
}

\maketitle

\begin{abstract}
Urban regions are complicated functional systems that are closely associated with and reshaped by human activities. The propagation of online geographic information-sharing platforms and mobile devices equipped with the Global Positioning System (GPS) greatly proliferates proximate sensing images taken near or on the ground at a close distance to urban targets. Studies leveraging proximate sensing imagery have demonstrated great potential to address the need for local data in the urban land-use analysis. This paper reviews and summarizes the state-of-the-art methods and publicly available data sets from proximate sensing to support land-use analysis. We identify several research problems in the perspective of examples to support the training of models and means of integrating diverse data sets. Our discussions highlight the challenges, strategies, and opportunities faced by the existing methods using proximate sensing imagery in urban land-use studies.


\end{abstract}

\section{Introduction}

Analysis of urban land-use enables researchers, practitioners, and administrators to understand city dynamics and to plan and respond to urban land-use needs. It also reveals human social activities in terms of locations and types in cities, which is closely related to human behaviors with respect to buildings, structures, and natural resources~\citep{wang08,Yuan11}. Applications such as urban planning, ecological management, and environment assessment~\citep{saynaj14} require the most updated knowledge of urban land-use. Conventionally, urban land-use information is obtained through field surveys, which is labor-intensive and time-consuming. The employment of proximate sensing data has demonstrated the potential of automatic, large-scale urban land-use analysis~\citep{leung09,Qiao20a} and thus attracted researchers from fields of computer science and geographic information systems~\citep{Qiao20b}.

Proximate sensing imagery, which refer to images of close-by objects and scenes~\citep{leung09}, complements the overhead imagery by providing information of objects from another perspective and brings completely disparate clues for urban land-use analysis. Urban land-use is closely related to human activities and demands more approximate means to scrutinize the cities~\citep{lefev17}. The crucial features associated with human activities are usually obscured from overhead imagery such as satellite images~\citep{karasov2019mapping}. For example, differentiating commercial (e.g., office buildings) and residential buildings (e.g., apartments) is a typical problem in urban land-use analysis and it is agreed in the research community that overhead imagery alone provides insufficient information for the aforementioned issue. Moreover, publicly available data that can be adopted as proximate sensing imagery are massive in volume. For example, over 300 million images are uploaded to Facebook every day~\citep{web7}, which enables the development of automatic, large-scale, data-driven approaches for urban land-use analysis.

\begin{figure}[!htb]
\centering
\includegraphics[width=4.5in]{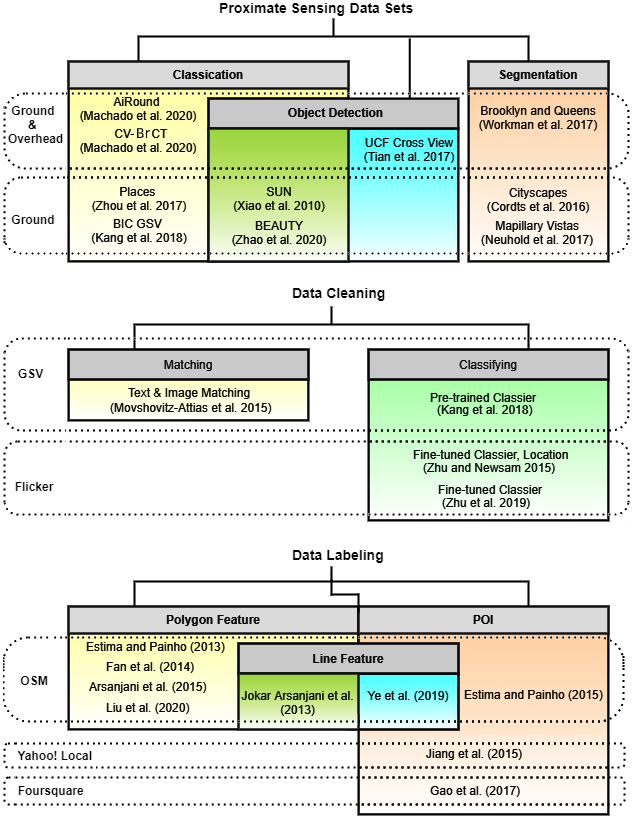}
\caption{Proximate sensing data sets and associated issues. \label{fig:data}}
\end{figure}

\begin{figure}[!htb]
\centering
\includegraphics[width=4.5in]{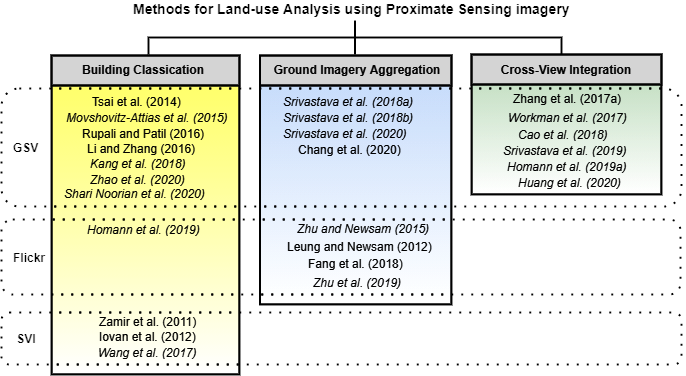}
\caption{Taxonomy of land-use analysis methods. SVI denotes the unspecified street view images. The ones in italic are methods that employ deep learning.  \label{fig:method}}
\end{figure}

This article is the first one that reviews the up-to-date studies on the employment of proximate sensing imagery for urban land-use analysis. The unique properties of proximate sensing imagery have motivated the development of novel methods, which necessitates a survey of data and methods to provide researchers a comprehensive review of the state-of-the-art. We categorize a diverse collection of emerging technological advancements on this topic and identifying technical challenges, existing solutions, and research opportunities. Throughout the literature, we observe challenges in two aspects: a myriad of data sets and technical obstacles. Discussions are hence assembled on these challenges. Figure~\ref{fig:data} illustrates the data sets and two associated issues: data cleaning and labeling. The AiRound,  CV-BrCT,  UCF  Cross  view,  and  Brooklyn and Queens data sets consist of ground-level and overhead images, and the rest only contains ground-level images. The majority data cleaning methods use classifiers to filter out incompatible examples and for land-use labeling, information from OpenStreetMap (OSM) serves as the primary reference. Figure~\ref{fig:method} shows the taxonomy of land-use analysis methods, which are grouped three categories: building classification, ground imagery aggregation, and cross-view integration. Google street view (GSV) serves as the major source of proximate sensing images in the land-use analysis. 

The remainder of this article is organized as follows. Section~\ref{sec:Data} summarizes the proximate sensing data for land-use analysis and presents the technical challenges in data cleaning and land-use example labeling. Section~\ref{sec:Methods} reviews the state-of-the-art methods from the perspectives of building classification, data aggregation, and cross-view land-use classification. Section~\ref{sec:Conclusion} summarizes this paper and highlights the opportunities for future research.

\section{Proximate sensing data and preprocessing \label{sec:Data}}
\subsection{Data Sets}

A vital source of proximate sensing imagery is the street view images provided by map service providers such as Google Street View, Apple Look Around, and Bing StreetSide. These services cover most major cities worldwide. In addition, companies such as Baidu, Tencent, Yandex, and Barikoi also provide regional street view images. Among these map service providers, GSV is the most influential geographical information service and was debuted in 2007. As of 2020, GSV has covered nearly 200 countries on four continents, which makes it an opportune data source for urban land-use analysis~\citep{wiki2}.

Another major source of proximate sensing imagery is the volunteer geographic information (VGI) made available via social media platforms such as OpenStreetMap, Instagram, Facebook, and Flickr. The affordability and portability of modern mobile devices rigged with cameras and GPS make every social media user a potential data provider~\citep{Terr21}. Consequentially, a large volume of images with GPS information has been created and continues to be updated every day. Such VGI data also contain annotations to assist urban land-use analysis~\citep{mahabir2020crowdsourcing, munoz2020openstreetmap}. \cite{anton16} reviewed VGI images for mapping land-use patterns and found that more than half of the collected images are helpful to extract the land-use related information. 
An emerging form of volunteered street view imagery, e.g., Mapillary and OpenStreetCam, provides a rich source of spatial geotagged street-level imagery along roads. The coverage of the VGI imagery, however, varies greatly within and across cities~\citep{Maha20}.

\begin{table}[!htb]
\caption{Proximate sensing imagery data sets. Image types include overhead (O), ground (G), multi-spectral (M). Cityscapes consists of images in fine and coarse resolutions. For the data sets that have object detection examples, the corresponding entries denote the number of annotated objects.\label{tb:Dataset}}
\centering
\begin{tabular}{p{2.4in}|lcl}
\hline
\multicolumn{1}{c|}{Data Set} & \# of Images & \# Class & Application \\ \hline\hline
Places \citep{zhou17}& 10,624,928 & 434 & Classification \\
\hline
BIC GSV \citep{kang18}& 19,658 & 8 & Classification \\
\hline
 & 1,165 (O) & & \\ 
AiRound \citep{machado2020airound} & 1,165 (G) & 11 & Classification \\ 
& 1,165 (M) & \\ \hline
CV-BrCT & 24,000 (O) & 9 & Classification \\ 
\citep{machado2020airound} & 24,000 (G) & \\ \hline
SUN & 131,072 & 908 & Classification \\ 
\citep{xiao10} & 313,884 & 4,479 & Obj. Detection \\ \hline
BEAUTY & 19,070 & 4 & Classification \\ 
\citep{zhao2020bounding} & 38,857 & 8 & Obj. Detection \\ \hline
UCF Cross View & 40,000 (O) & - & Obj. Detection\\
\citep{tian17} & 15,000 (G) & & \\ \hline
Brooklyn and Queens & 53,649 (O) & 206 & Segmentation \\ 
\citep{workm17} & 177,930 (G) & \\ \hline
Cityscapes & 5,000 (fine) & 30 & Segmentation \\ 
\citep{cordt16} & 20,000 (coarse) & \\ \hline
Mapillary Vistas & 25,000 & 152 & Segmentation \\ 
\citep{neuho17} & & & \\ \hline
\end{tabular}
\end{table}

Table~\ref{tb:Dataset} summarizes the data sets adopted in the previous studies. To our best knowledge, there is no widely adopted benchmark proximate sensing imagery data set for urban land-use analysis yet. The AiRound, CV-BrCT, UCF Cross view, and Brooklyn and Queens data sets include both proximate sensing data and overhead imagery; the rest contain only proximate sensing images. Among these data sets, BIC GSV, AiRound, CV-BrCT, and Brooklyn and Queens data sets are designed specifically for the task of urban land-use related classification. UCF Cross View data set can be used to match overhead and proximate images. Places, SUN, Cityscapes, and Mapillary Vistas data sets are relatively large-scale, and a portion of the data and annotation information can be leveraged for urban land-use classification.

\subsection{Data cleaning}

A challenge faced by employing the existing data sets is inconsistency, which necessitates data cleaning. The cleaning and refinement of proximate sensing images is a non-negligible problem. Proximate sensing data vary greatly in contrast to data sets for many remote sensing applications, and the major issue is three-fold. 
\begin{enumerate}
\item {\it Only a portion of the images captured from the ground perspective includes spatial, contextual information for urban land-use analysis.} 
The geo-tagged images available in online services such as Flickr and Facebook contain a large number of selfies, photographs of food, pets, and other contents that provide little help in understanding urban structures and land-use. 

\item {\it There exists a disconnection between the contents of an image and its geographic coordinates.} 
Images are often captured views at a distance from the shooting point of the photographer. The geographic coordinates automatically embedded in these images reflect the location of the photo taker instead of the objects captured in the image. The images illustrate buildings or structures that are outside of the current land-use functional unit. 

\item {\it Useful information of the image sets after removing irrelevant instances is limited for achieving high accuracy with satisfactory robustness.} 
Objects in images that provide hints of land-use may be insignificant due to the small size or off the center of the image. After all, these images are not taken intentionally for land-use classification, which makes data cleaning a crucial component in the process of land-use analysis. 
\end{enumerate}

To sift usable data from street view images, \cite{movsh15} constructed a database of manually identified business entities that are presented by location and textual information. The same description of unlabeled street view images was generated. The business entity was assigned to a street view image if the distance between the entity and the image is less than one street block. Images with irrelevant information or taken from a distance were discarded. 
\cite{zhu15} employed polygon outlines and classifiers to clean VGI data. Flickr images that are outside of the extracted polygonal regions are removed. A search strategy was used for data augmentation to ease the imbalance among classes of the training data. 

An alternative means for data cleaning is applying pre-trained deep network models. 
\cite{kang18} adopted the VGG16~\citep{simon14} model fine-tuned on Places2 data set~\citep{zhou16}. A large number of training examples in Places2 data set and the overlapping of Places2 data and proximate sensing data make it a proper source to fine-tune the VGG16 model for land-use classification. The fine-tuned model was used to decide if an image is relevant to urban land-use. 
\cite{zhu19} developed an online training method for data cleaning. To create a relatively large data set for fine-grained urban land-use classification, both Flickr and Google Images were used. The online adaptive training was implemented following the intuition that if the Softmax scores of an image are evenly distributed among all categories, this image is likely to be confusing and irrelevant to land-use analysis. Images with strongly skewed prediction scores benefit the development of the model. Similar to negative mining~\citep{Yuan02,you15,shriv16}, samples that result in a low probability value are discarded.

Table~\ref{tb:Cleaning} summarizes the existing data cleaning methods. Most of the efforts are based on applying a classifier to identify suitable (or unsuitable) instances. Despite the effectiveness of removing loosely related instances, a gap between image content and land-use types still exists. The development of novel methods that automatically select the most representative images or preclude less informative ones is still of great importance.

\begin{table}[!htb]
\caption{Data cleaning strategy. \label{tb:Cleaning}}
\begin{center}
\begin{tabular}{l|c|l}
\hline
\multicolumn{1}{c|}{Method} & Data set & \multicolumn{1}{c}{Strategy} \\ \hline\hline
\cite{movsh15} & GSV & Text \& Image Matching \\ 
\cite{kang18} & GSV & Pre-trained Classifier \\
\cite{zhu15} & Flickr & Fine-tuned Classifier, Location \\
\cite{zhu19} & Flickr & Fine-tuned Classifier, Location \\
\hline
\end{tabular}
\end{center}
\end{table}

\subsection{Land-use labeling}

Proximate sensing requires labels for buildings and urban functional regions for land-use analysis. The labels of land-use are often not readily available for the images, which poses a great challenge in research and deployment. OSM tags and the Point of Interest (POI) have been used in studies to derive land-use labels and annotate proximate sensing images~\citep{Varg20}. The OSM database consists of several sub-sets: points, places, roads, waterways, railways, buildings, land-use, and natural areas. OSM and map service providers such as Yahoo! Local~\citep{jiang15}, Foursquare~\citep{gao17}, Google Places, and \cite{web3} also provide geo-tagged POI data. However, the quality of OSM tags and POIs is usually undermined due to limited regularization and censorship. Hence, studies have been conducted to understand the usage of OSM tags and POIs. 

An early study of deriving labels from OSM data was conducted by~\cite{hakla08}. The study found that the OSM tags are suitable for land-use analysis with an accuracy of 80\% comparing to the existing survey data. \cite{estim13} employed OSM tags for land-use classification and achieved an accuracy of 76\%. \cite{fan14} asserted that OSM tags contain a vast amount of building information and the size and shape of building footprints also provide clues to the function of buildings. \cite{arsan15} evaluated OSM tags in four metropolitan areas in Germany and the Global Monitoring for Environment and Security Urban Atlas data (GMESUA)~\citep{Urban} data set was used as the reference. \cite{Font17} assessed the OpenStreetMap for the creation of reference databases in the evaluation of land-use/land cover maps. The study concluded that a small portion of the OSM based reference data set requires photointerpretation of high-resolution imagery.  

Besides OSM data, POIs have been heavily used to generate urban land-use labels for proximate sensing imagery. \cite{estim15} explored the POI data extracted from OSM in an area of Continental Portugal. The experiments demonstrated that among the 26,191 POI examples, the agreement rate to the official land-use data is 78\%. \cite{jiang15} devised a set of rules to compare POIs and map POIs provided by different data sets. The POIs were then aggregated with retail employment data. \cite{gao17} developed a statistical framework based on the latent Dirichlet allocation topic model to discover urban functional regions. The study concludes that consociating the spatial pattern distribution of POIs helps extract urban functional regions.

Combining OSM tags and POIs for data labeling has also been investigated. \cite{jokar13} developed a hierarchical GIS-based decision tree to generate the land-use map from OSM point, line, and polygon features. \cite{ye19} fused OSM tags, POIs, and satellite images and proposed a Hierarchical Determination method that extracts roads from OSM to generate functional units.  
\cite{liu2020annual} randomly sampled points in the OSM polygons and used the OSM tags as the label for these points. This method built a semi-automatic framework to map urban land-use using OSM data.

Despite the great success of using OSM and POI data for label generation, online services that rely on voluntary contributors face issues of great inconsistency and errors. The major blemishes of OSM tags include misaligned tags with the images and buildings as well as missing annotations for buildings.  To address these issues, \cite{varga19} developed a tag correction method based on Markov Random Field and Convolutional Neural Network (CNN). A probability map was constructed from the correlation between tags and buildings and a CNN model was trained from building shapes to assign labels for buildings without a label. 

Table~\ref{tb:Label} summarizes the existing land-use labeling methods. Among all the supplementary data source, OSM serves as a major source for automatic land-use annotations. These studies demonstrated that OSM tags and POI data provide valuable information for land-use labeling of proximate sensing imagery. However, OSM, as well as other similar service providers, allow users to define their labels (or tags). This enables the flexibility and adaptability of tagging but increases inconsistency and bewilderment of using the tagged data for automatic land-use labeling. Aligning the tags and the land-use types has not yet fully studied. Label extraction, alignment, sorting, and refinement are still subjective and obscure. The development of automatic methods for land-use labeling of the proximate sensing imagery is needed.

\begin{table}[!htb]
\caption{Land-use labeling. The rows with multiple \# Class values denote that the classification was performed in multiple levels following a coarse to fine manner. \label{tb:Label}}
\centering
{\small
\begin{tabular}{l|c|c|l}
\hline
\multicolumn{1}{c|}{Method} & Source of Label & \# Class & \multicolumn{1}{c}{Feature} \\ \hline\hline
\cite{estim13} & OSM & 5, 15, 44 & Polygon  \\
\cite{fan14} & OSM & 6 & Polygon  \\
\cite{arsan15} & OSM & 15 & Polygon  \\
\cite{liu2020annual} & OSM & 16 & Polygon  \\
\cite{jokar13} & OSM & 2, 4, 15 & POI, Line, Polygon \\
\cite{ye19} & OSM & 10 & POI, Line  \\
\cite{estim15} & OSM & 5, 15, 44 & POI \\
\cite{jiang15} & Yahoo! Local & 14 & POI \\
\cite{gao17} & Foursquare & \multicolumn{1}{c|}{-} & POI \\
\hline
\end{tabular}
}
\end{table}

\section{Methods for land-use analysis}\label{sec:Methods}

\subsection{Building classification}

Differentiating building usage from overhead images is ill-posed in urban settings due to the uncertainty of correspondence between roof tops and surroundings of a building to its actual usage. Proximate sensing images enable researchers to integrate building side views, texture, and decorations (e.g., signs and sculptures) to more accurately decide the building usage. 

An early exploration of associating street view images with building functions was conducted by \cite{zamir11}. In this study, a set of 129,000 street view images and textual information were used to identify commercial entities. The list of businesses was generated from services such as Yellow Page and the text information detected from the street view images are matched to the business entities using Levenshtein distance. The experiments achieved an overall accuracy of 70\%. 
\cite{iovan12} used scale-invariant feature transform (SIFT) descriptors randomly sampled from each image to create a visual dictionary. Bag of Words (BoW) model~\citep{zhang10} and Bag Of Statistical Sampling Analysis (BOSSA) model~\citep{avila11} were applied to generate images signatures. Using a kernel Support Vector Machine (SVM), the method classifies urban structures such as shops, porches, etc. 
\cite{tsai14} employed OpponentSIFT~\citep{van09} as image features and created a codebook using clusters of BoW features. The recognition was conducted using a distributional clustering. A similar strategy was implemented by \cite{rupal16}, in which SIFT descriptor and clustering were used. The proposed two-phase framework recognizes the on-premise signs of business entities using street view images. 
\cite{li16} used the GSV images of New York City to differentiate single-family buildings, multi-family buildings, and non-residential buildings. Feature descriptors such as GIST, HoG, and SITF-Fisher were implemented for classification. It was demonstrated that SIFT-Fisher descriptor achieved the best accuracy of 91.82\% on classifying residential and non-residential buildings. 

Besides the aforementioned feature engineering methods, deep networks, especially CNN models pre-trained on large scale data sets, have been broadly used for building function classification. \cite{movsh15} created a large training data set using an ontology-based labeling method, which were used to learn multi-label, fine-grained storefronts. A CNN model based on GoogLeNet was trained with ImageNet~\citep{deng09} and fine-tuned using street view images. 
\cite{wang17a} employed AlexNet~\citep{krizh12} to classify stores from street view images. 
\cite{kang18} used a CNN model trained with Place2 to filter out images irrelevant to buildings and employed pre-trained deep networks including AlexNet, ResNet18, ResNet34, and VGG16 for building classification. 
\cite{hoffm19b} performed a five-class classification using geo-tagged images from Flickr and supplementary building polygons from OSM. A spatial nearest neighbor classifier was developed to assign images to buildings. A VGG16 model pre-trained with ImageNet was adopted for feature extraction and a logistic regression classifier trained using SAGA optimizer~\citep{defaz14} was applied to make the final prediction. 

Object detection has been employed for building classification. \cite{hoffm19} used a ResNet50 based the Single Shot MultiBox Detector~\citep{liu16} trained with the COCO data set~\citep{lin14} to detect the frequently appeared objects in social media images. The rasterization was performed by counting the detected objects, the mutual information between the object frequency and the function of the nearby buildings was computed. The study found a strong correlation between the object counts in social media images and the building functions. 
\cite{zhao2020bounding} devised a `Detector-Encoder-Classifier' network to detect buildings in GSV images using object detectors~\citep{ren15,cai2018}, which were fed into a Recurrent Neural Network (RNN) for urban land-use classification. A recent study by~\cite{sharifi2020detecting} implemented a framework to classify the retail storefronts using GSV images. YOLOv3~\citep{redmon2018yolov3} was applied to detect the storefronts, then ResNet pre-trained using Places365 data set was used to further perform the classification. 

Table~\ref{tb:Building} summarizes the methods for building classification. Among the available imagery data, street view images, especially GSV images, serve as a major source of proximate sensing data for building classification. Besides using conventional image features, studies conducted by~\cite{movsh15}, \cite{wang17a} and \cite{hoffm19b} adopted CNN models that follow an end-to-end design and, hence, integrate feature extraction with classification. Multi-functional buildings (e.g., apartment buildings with restaurants on the ground floor) pose greater difficulty in comparison to the single functional ones, which appear often in large metropolitan and dense urban areas. The development of multi-label classification could be responsive to such a unique problem. In addition, leveraging interior photographs demonstrated the potential for fine-grained building classification but calls for further exploration.

\begin{table}[!htb]
\caption{Building classification methods. SVI denotes unspecified street view images; GSV denotes Google street view images; Deep represents deep features. \label{tb:Building}}
\centering
{\small
\begin{tabular}{l|l|c|l}
\hline
\multicolumn{1}{c|}{Method} & Data & \# Class & Classifier (Feature)\\ 
\hline\hline
\cite{zamir11} & & 2 & Levenshtein Dist. (Text, Gabor)\\
\cite{iovan12} & SVI & 4 & SVM (SIFT, BoW, BOSSA) \\
\cite{wang17a} & & 8 & AlexNet (Deep)\\ 
\hline
\cite{tsai14} & & 62 & Thresholding (SIFT)\\
\cite{movsh15} & & 208 & GoogLeNet (Deep)\\ 
\cite{rupal16} & & 62 & Thresholding (SIFT)\\ 
\cite{li16} & GSV & 4 & SVM (GIST, HOG, SIFT) \\
\cite{kang18} & & 8 & AlexNet, ResNet, VGG (Deep)\\
\cite{zhao2020bounding} & & 4 & Cascaded R-CNN, RNN (Deep)\\
\cite{sharifi2020detecting} & & 24 & YOLOv3, ResNet (Deep )\\
\hline
\cite{hoffm19b}& Flickr & 5 & Logistic Regression (Deep )\\ 
\hline
\end{tabular}
}
\end{table}

\subsection{Aggregation of proximate sensing imagery}

The proximate sensing imagery are largely diverse in contents, perspectives, and field of views. Imagery data from online social networks and mapping services facilitated the practicability of aggregating various proximate sensing images to reform urban land-use classification. \cite{leung12} explored the Flickr images of two university campuses. Geo-tagged images are grouped based on the spatial locations, contributor, and acquisition time. Text annotations were used as auxiliary data for training an SVM classifier. 
Images taken from the building interior and surrounding areas provide additional clues of human activities and, hence, can be used as auxiliary data sources. \cite{zhu15} extended the method by differentiating indoor and outdoor images with a classifier trained with the SUN data set~\citep{xiao10} and extracting semantic features using a pre-trained CNN model. The aggregation was achieved by majority voting. 
\cite{fang18} integrated OSM data with geo-tagged images from social networks for urban land-use classification. The urban space is divided using the hierarchical urban street networks. Object Bank (OB)~\citep{li10} was used to extract features and predict labels to individual image. The land-use type of each parcel is generated by weighted sum the image classes within within the parcel. 
\cite{chang2020mapping} leveraged the semantic segmentation of GSV images to construct a representation for urban parcels. The features from GSV, Luojia-1, Sentinel-2A, and Baidu POIs were integrated. A random tree was implemented for classification. 

To enrich the training data for the employment of deep networks, images from multiple platforms are often used. \cite{trace17} employed VGI images from Flickr, Panoramio, Geograph, and Instagram. A CNN trained with the aggregated data sets was fine-tuned for land-use classification. 
\cite{zhu19} built a large scale, fine-grained land-use data set that include images from Flickr and Google Images. A two-stream model was developed for object recognition and scene recognition. The object stream was a CNN model pre-trained with ImageNet and the scene stream is another CNN model pre-trained with the Places365 data set~\citep{zhou14}. 
\cite{srivas18} adopted CNN models for the task of multi-label building function classification. The building labels were derived from Addresses and Buildings Databases~\citep{web1}. Features of three street view images of different field of views at each street location were extracted using a pre-trained VGG16 model for classification. It was demonstrated that aggregated network outperforms the uni-modal network and the vector stacking method. Followup studies \citep{srivas18a,sriva18c} demonstrted that using multiple images at the same location improves the accuracy of land-use classification. 

Table~\ref{tb:Agg} summarizes the methods that aggregate multiple approximate sensing images for urban land-use classification. Besides the conventional semantic features, BoW and OB are used for feature extraction. The dominant strategies for aggregation include feature level concatenation and averaging and decision level majority voting. The key motivation is that each image represents only a partial view of the land unit; hence, aggregating multiple views from different perspectives results in an informed decision. Apart from multi-perspective images, \cite{leung12} leveraged text information from Flickr as an auxiliary source of information, which demonstrated the feasibility of integrating dramatically different information for improved performance.

\begin{table*}[!htb]
\caption{Methods of land-use classification that combine proximate sensing images. GSV denotes Google Street View images; GI denotes Google Images; Deep represents deep features; Ave. denotes the average aggregator; Con. stands for concatenation. \label{tb:Agg}}
\centering
{\small
\begin{tabular}{p{1.6in}|l|c|c|c|c|c}
\hline
 & Source &\# of & \multicolumn{2}{c|}{Feature Level Fusion} & \multicolumn{2}{c}{Decision Level Fusion} \\ \cline{4-7}
\multicolumn{1}{c|}{Method} & of Data & Class & Feature & Strategy & Classifier & Strategy \\ \hline\hline
\cite{fang18} & Flickr & 5 & OB & \multicolumn{1}{c|}{-} & SVM & Voting \\ 
\cite{zhu15} & Flickr & 8 & Deep & \multicolumn{1}{c|}{-} & SVM & Voting \\ 
\cite{zhu19} & Flickr/GI & 45 & Deep & \multicolumn{1}{c|}{-} & ResNet & Ave. \\ 
\cite{leung12} & Flickr & 3 & BoW & Ave & SVM & \multicolumn{1}{c}{-} \\ 
\cite{srivas18a} & GSV & 13 & Deep & Ave & SVM, MLP & Voting \\ 
\cite{srivas18} & GSV & 9 & Deep & Con. & VGG16 & \multicolumn{1}{c}{-} \\ 
\cite{sriva18c} & GSV & 16 & Deep & Ave/Max & VGG & \multicolumn{1}{c}{-} \\
\cite{chang2020mapping} & GSV & 5 & Numeric & Con. & - & \multicolumn{1}{c}{-} \\ \hline
\end{tabular}
}
\end{table*}

\subsection{Integrating imagery of different perspectives}
An intuitive way to integrate images of different perspectives is constructing a pixel-level land-use map. 
\cite{workm17} combined overhead and proximate images for land-use, building function classification, and building age estimation. The data set consists of GSV, Bing Map, and official city planning information. Two pre-trained VGG-16 models were used to extract features from street view images as well as overhead images. Hypercolumn was extracted from the feature maps using PixelNet~\citep{bansa17}. It was demonstrated that the top-1 accuracy of land-use classification by combining overhead and proximate sensing images achieved an improvement of 11.2\%. 
\cite{cao18} extracted the features of street view images using PlacesCNN and used Nadaraya-Watson kernel regression for spatial interpolation. After constructing the ground feature map, a SegNet~\citep{badri17} based network is used to integrate the overhead imagery and ground feature map and perform the land-use classification. The proposed network contains two VGG16 based encoders that produce a pixel-level urban land-use map with a decoder. 
\cite{feng18} developed a multi-view CNN for pixel-level segmentation. In this network, lower-order potentials were used for processing overhead images and higher-order potentials were sued for proximate sensing images. Feature stacking was used to achieve fusion of proximate sensing and overhead images. 

An alternative strategy is deciding land-use types for each parcel. 
\cite{zhang17} developed an urban land-use data set including overhead LiDAR, high-resolution orthoimagery (HRO), GSV, and parcel data. The method assumes that the existence of text in the street view images is an essential indicator to differentiate residential and non-residential buildings, which was achieved by classifying GSV images for text detection. 
The classification accuracy achieved a improvement 29.4\% in classifying mix residential buildings.
\cite{huang20} applied pre-trained DeepLabV3+~\citep{chen2018encoder} and ResNet-50~\citep{he2016deep} on satellite and GSV imagery to learn land cover proportion and scene category of each parcel. Features extracted from building footprint, POI, and check-in data were fed into an XGBoost classifier for urban land-use classification.

Research has been conducted to associating proximate sensing images to urban objects or buildings for land-use mapping. 
\cite{sriva19c} associated the GSV images with urban-object footprints extracted from OSM. The proposed method integrated overhead and proximate sensing images with a two-stream CNN model: a patch-based classification~\citep{penat15} for extracting features from overhead images and a Siamese model~\citep{bromle94} for proximate sensing images. It was  demonstrated that multi-model CNN models outperform uni-modal CNN models. The overall accuracy was at 75.07\%. 
\cite{hoffm19a} use the building function information provided by OSM and associate it with corresponding GSV and overhead images. Two fusion strategies were implemented: geometric feature fusion and decision fusion. Geometric feature fusion follows the two-stream model and the decision-level fusion model is based on model blending and stacking. The experiments demonstrated the decision fusion outperformed the feature fusion model. 

\begin{table*}[!htb]
\caption{Methods of land-use classification that combine data of cross-view modalities. Prox. and Over. denotes proximate and overhead data, respectively; Strat. stands for strategies used in the corresponding method; GSV denotes Google Street View images; Deep represents deep features; Ave. denotes the average aggregator; Con. stands for concatenation. \label{tb:Fusion}} 
\centering
{\small
\begin{tabular}{p{1.45in}|l|l|c|c|c|c|c}
\hline
& Prox. & Overhead & \# of & \multicolumn{2}{c|}{Feature Fusion} & \multicolumn{2}{c}{Decision Fusion} \\ \cline{5-8}
\multicolumn{1}{c|}{Method} & Data & Data & Class & Feature & Strat. & Classifier & Strat. \\ \hline\hline
\cite{zhang17a} & GSV & LiDAR/Sat. & 7 & Numeric & Con. & RF & \multicolumn{1}{c}{-} \\ 
\cite{workm17} & GSV & Satellite & 206 & Deep & Con. & MLP & \multicolumn{1}{c}{-} \\
\cite{workm17} & GSV & Satellite & 11 & Deep & Con. & MLP & \multicolumn{1}{c}{-} \\
\cite{cao18a} & GSV & Satellite & 13 & Deep & Con. & SegNet & \multicolumn{1}{c}{-} \\ 
\cite{sriva19c} & GSV & Satellite & 16 & Deep & Con. & VGG & \multicolumn{1}{c}{-} \\ 
\cite{hoffm19a} & GSV & Satellite & 4 & Deep & A/C & VGG & A/C \\
\cite{huang20} & GSV & Satellite & 8 & Deep & Con. & XGBoost & - \\ \hline
\end{tabular}
}
\end{table*}

Table~\ref{tb:Fusion} summarizes the methods that integrate images acquired from different perspectives (i.e., proximate sensing images and remote sensing images). Most of the methods extract and combine features from street view and satellite images via concatenation, which are then used as inputs for a classifier. Recently, an attempt of employing both feature level and decision level fusion has been performed~\citep{hoffm19a}. The decision fusion was achieved by tallying class scores. The advantage appears to be incremental and needs to be confirmed. Besides satellite images, LiDAR data were also used; yet, the results remain limited.

\section{Discussion}

Table~\ref{tb:accuracyComparison} summarizes the methods, data set used, properties of the applications (number of classes and number of images used in training), and the accuracy. The column ‘number of Images’ reports the number of proximate sensing images used in the studies. The numbers in italic are precision instead of accuracy. The results by \cite{li16} report two cases: residential building vs non-residential building and single-family building vs multi-family building. Hence, the accuracy are reported separately. In the studies conducted by \cite{workm17}, the accuracy include the combination of number of classes and number of examples used. In case of \cite{cao18a} and \cite{sriva19c}, two data sets from different locations were used in evaluation, which produced different results. 

In this table, we organize the methods according to the problems they address and the data sets used. Despite three data sets were used, the settings of experimental evaluation vary greatly. Hence, it is difficult to draw conclusions on the state-of-the-art performance. It is clear that when a large number of classes exists (e.g., 20 or more classes), the average accuracy is inferior to the cases where a smaller number of classes needs to be differentiated. From the perspective of applications, the average accuracy of building classification is 71.60\% and the median accuracy is 69.43\%. For the ground-view aggregation, the average accuracy and median accuracy are 65.31\% and 69.05\%. For cross-view integration, the average accuracy and median accuracy are 68.74\% and 74.87\%.

\begin{table}[!htb]
\caption{Accuracy/precision (\%) comparison of land-use classification methods using proximate sensing images. ``Number of Images" only includes the number of proximate sensing images used in corresponding researches. SVI denotes unspecified street view images. GSV denotes Google street view images. GI stands for Google Images. The entries Enclosed in parentheses indicate precision is reported instead of accuracy.
\label{tb:accuracy}}
\centering
{\small
\begin{tabular}{c|l|c|c|c|c}
\hline
 & & Source &\# of & \# of & Acc. \\ 
Problem & \multicolumn{1}{c|}{Method} & of Data & Classes & \ Images & (Prec.) \\\hline\hline
&\cite{zamir11} & SVI & 2 & 129,000 & 70.00 \\
&\cite{iovan12} & SVI & 4 & 1,516 & 76.21 \\
&\cite{wang17a} & SVI & 8 & 4,636 & 93.60 \\ 
&\cite{tsai14} & GSV & 62 & 4,649 & (68.60) \\
Building &\cite{movsh15} & GSV & 208 & 1,300,000 & (63.00) \\ 
Classification &\cite{rupal16} & GSV & 62 & 4,649 & (68.86) \\ 
&\cite{li16} & GSV & 2 & 1,048 & 91.82 \\
&\cite{li16} &GSV & 2 & 1,048 & 74.30\\
&\cite{kang18} & GSV & 8 & 19,658 & (59.00)\\
&\cite{zhao2020bounding} & GSV & 4 & 19,070 & (81.81) \\
&\cite{sharifi2020detecting} & GSV & 24 & 1,200 & 45.01 \\
&\cite{hoffm19b}& Flickr & 5 & 343,600 &(67.00) \\ \hline
&\cite{fang18} & Flickr & 5 & 24,835 & 76.50 \\ 
&\cite{zhu15} & Flickr & 8 & 37,784 & 76.00 \\ 
Ground-View&\cite{zhu19} & Flickr/GI & 45 & 58,418 & 49.54 \\ 
Aggregation &\cite{srivas18a} & GSV & 13 & 3,4261 & 69.05 \\ 
&\cite{srivas18} & GSV & 9 & - & 44.41 \\ 
&\cite{sriva18c} & GSV & 16 & 4,4957 & 62.52 \\
&\cite{chang2020mapping} & GSV & 5 & - & 79.13\\ \hline
&\cite{workm17} & GSV & 206 & 139,327 & 44.88 \\
&\cite{workm17} & GSV& 11 & 139,327 & 77.40\\ 
 &\cite{workm17} & GSV & 206 & 38,603 &34.13\\
 &\cite{workm17} & GSV&11 & 38,603 & 70.55\\ 
Cross-View&\cite{zhang17a} & GSV & 7 & - & 77.50 \\
Integration& \cite{cao18a} & GSV & 13 & 139,327 & 78.10 \\ 
& \cite{cao18a} & GSV& 13 & 38,603 & 74.87 \\
& \cite{sriva19c} & GSV & 16 & 44,957 & 73.44 \\
& \cite{sriva19c} & GSV& 16 & 9,908 & 75.07 \\
&\cite{hoffm19a} & GSV & 4 & 225,036 & (76.00) \\
&\cite{huang20} & GSV & 8 & 660,000 & 74.20 \\\hline
\end{tabular}\label{tb:accuracyComparison}
}
\end{table}

The collection and sharing proximate sensing data involves ethical issues such as privacy and trust of anonyms. With a wide application of smartphones and dash-cams, images with embedded metadata are collected automatically, which are shared via online social networks. In addition, sensitive personal information such as license plates and biometric data are captured without a consent. Enterprises have developed policy and measures to address the privacy issues by blurring pedestrian’s face and the license plate numbers and allowing users submit requests to remove or obfuscate personal information, e.g., face and home view. A similar practice is implemented in platforms such as OpenStreetCam. Another aspect is related to the commitment and professionalism of the contributors. The open platforms that allow users to upload images and videos voluntarily face challenges of ensuring data quality. Users contribute to data without risking the irrevocable consequences. This necessitates the data cleaning process for using imagery data from open platforms as well as studies to evaluate the quality of data (Mahabir et al. 2020).

As illustrated in Table~\ref{tb:accuracyComparison}, the data used in different methods vary greatly. This makes the comparison and an understanding of the state-of-the-art difficult. On one hand, this is because the rule for defining land-use classes is not unified, so different researcher may choose different standard to format their data. Previous studies majorly adopt two ways to build and label their proximate sensing data. One is manually define the land use classes in a controlled environment~\citep{zhu15} or manually select the proximate sensing images that fit their pre-designed land-use labels~\citep{kang18, zhao2020bounding}. The other way is using the information provided by government or official institutions to guide the formatting of their data. For example, \cite{workm17} referred to the documentation of New York City Department of City Planning to label their data sets; \cite{zhu19} adopted the Land Based Classification Standards to generate the labeling information. 

On the other hand, the fact that there are no large-scale, annotated, public available proximate sensing data sets forces the researchers to build their own data sets, thus gives rise to the inconsistency of labeling. Although proximate images are large in amount, the amount of annotated data remains limited. Vast amount of proximate sensing data need to be automatically converted into training examples to support deep learning as well as consistent comparison. 

Also, proximate sensing images, especially volunteer images are biased to most prosperous areas of the cities, such as landmarks and attractions. Thus, the proximate sensing images data sets are often imbalanced. This phenomenon can be alleviated by gathering more data from a certain location~\citep{srivas18a, srivas18} or make use of supplemental data to augment the data sets~\citep{zhu15, zhu19}. For the pixel-level land-use classification, there exists a gap between consistent land-use map and sparse distributed proximate sensing images. Kernel regression and density estimation are used to convert the imbalanced proximate sensing features into a dense feature map, and the dense feature will be further align with the pixel-level land-use map~\citep{workm17, feng18, cao18}.

\section{Conclusion}\label{sec:Conclusion}

The urban landscape is formed by government planning and reshaped by the activities of the inhabitants. The identification of the functionalities of urban space is by nature tackling the ``problems of organized complexity'' \citep{fulle17}. The emergence of proximate sensing imagery has spurred many inspiring studies for better urban land-use analysis.

In this paper, we present the annotated data sets applicable to urban land-use analysis. The paper highlights problems of the proximate sensing imagery, i.e., data cleaning and labeling, and summarizes the methods to circumvent these problems. Due to the voluntary nature of most proximate sensing data sets, data quality and annotation availability are pressing issues. Several data refinement techniques were developed such as leveraging text, location, polygonal outline information to remove unusable data. Alternatively, using a pre-trained and fine-tuned model to filter out the irrelevant images is an acceptable approach. To automate the process of generating land-use annotations (labels), OSM tagging, and POI information have been employed and demonstrated effectiveness in the form of auxiliary information for urban land-use labeling.

Furthermore, we categorize the existing methods for land-use classification using proximate sensing imagery based on their underlying ideas. In particular, in early studies, conventional image features such as SIFT, HOG, GIST, and BoW are combined with classifiers to predict building functions; in the most recent studies, deep learning frameworks are employed, which enabled the deployment of end-to-end learning and further improved the accuracy. As surplus and complementary proximate sensing data are available, methods to integrate such information have been developed. To aggregate the multiple proximate sensing imagery of the same region, image-level features or decision-level features are extracted and integrated to form a consolidated input to the classifier. To leverage complementary overhead and proximate sensing imagery, feature stacking and feature fusion strategies are used to jointly learn the multi-modal feature representations, and decision fusion was adopted to achieve classification in a part of the studies. The studies demonstrated the effectiveness of leveraging proximate sensing imagery for urban land-use analysis, especially for differentiating residential and commercial entities and fine-grained urban land-use classification.

Despite the advancement demonstrated by many studies, leveraging proximate sensing imagery for urban land-use analysis remains an immature research area. To date, well-annotated data set suitable for such studies is still very limited. The demand for well designed, high-quality benchmark data is a pressing aspect for the continuation of this research field. Although supplementary data such as OSM and POI have exhibited promising value to automatic urban land-use annotation, refinement, sorting, and alignment of labels remain a non-trivial task. Thus, developing unsupervised or semi-supervised learning strategies could bring signification benefit in future studies. In addition to developing annotation-independent frameworks, introducing supplementary data, e.g., Google Images filtered by corresponding key words, or the land-use related entries in large-scale scene data sets could help to boost the model performance for land-use classification tasks. Moreover, the majority of current proximate sensing methods only use base benchmark CNN backbones. As proximate sensing images share mutual characteristics with some data sets in computer vision domain (e.g. scene recognition data sets),   adopting advanced network design strategies such as multi-scale frameworks or attention modules is promising to enhance the model performance on proximate sensing data. Besides, as land-use classification is not considered as a real-time task, the methods that require large models or intensive computation such as model integration and multi-modal techniques are appropriate for deploying proximate sensing researches. Thus, developing models aggregating information from multiple images and data of different perspectives remains a prospective strategy. 

With specific focus on land-use classification using proximate sensing images, this survey provides an overview of the recent progress in related literature. We hope this effort will drive further interest in the both geography and vision community to leverage the potential of proximate sensing images and improve on their current limitations.

\section*{Data availability}
Data sharing is not applicable to this article as no new data were created or analyzed in this study.

\bibliographystyle{tGIS}
\bibliography{ref}

\begin{thebibliography}{92}
\providecommand{\natexlab}[1]{#1}

\bibitem[\protect\citeauthoryear{Antoniou {\itshape{et~al.}}}{2016}]{anton16}
Antoniou, V., {\itshape et~al.}, 2016. Investigating the feasibility of
  geo-tagged photographs as sources of land cover input data. {\itshape {ISPRS
  International Journal of Geo-Information}}, 5 (5), 64.

\bibitem[\protect\citeauthoryear{Arsanjani {\itshape{et~al.}}}{2015}]{arsan15}
Arsanjani, J.J., {\itshape et~al.}, 2015. 2. {\itshape {In}}: {\itshape Quality
  assessment of the contributed land use information from {OpenStreetMap}
  versus authoritative datasets}.,  37--58  Springer, Cham.

\bibitem[\protect\citeauthoryear{{ATTOM Data Solutions}}{2020}]{web3}
{ATTOM Data Solutions}, 2020. {Points of interest data [online]}.  {Available
  from https://www.attomdata.com/data/neighborhood-data/points-interest-data/
  [Accessed November 2020]}.

\bibitem[\protect\citeauthoryear{Avila {\itshape{et~al.}}}{2011}]{avila11}
Avila, S., {\itshape et~al.}, 2011. {Bossa: Extended bow formalism for image
  classification}. {\itshape {In}}:  {\itshape {18th IEEE International
  Conference on Image Processing}},  2909--2912.

\bibitem[\protect\citeauthoryear{Badrinarayanan
  {\itshape{et~al.}}}{2017}]{badri17}
Badrinarayanan, V., Kendall, A., and Cipolla, R., 2017. {Segnet: A deep
  convolutional encoder-decoder architecture for image segmentation}. {\itshape
  {IEEE Transactions on Pattern Analysis and Machine Intelligence}}, 39 (12),
  2481--2495.

\bibitem[\protect\citeauthoryear{Bansal {\itshape{et~al.}}}{2017}]{bansa17}
Bansal, A., {\itshape et~al.}, 2017. {Pixelnet: Representation of the pixels,
  by the pixels, and for the pixels}. {\itshape arXiv preprint
  arXiv:1702.06506}.

\bibitem[\protect\citeauthoryear{Bromley {\itshape{et~al.}}}{1994}]{bromle94}
Bromley, J., {\itshape et~al.}, 1994. Signature verification using a ``Siamese"
  time delay neural network. {\itshape {In}}:  {\itshape {Advances in Neural
  Information Processing Systems}},  737--744.

\bibitem[\protect\citeauthoryear{Cai and Vasconcelos}{2018}]{cai2018}
Cai, Z. and Vasconcelos, N., 2018. Cascade R-CNN: Delving into high quality
  object detection. {\itshape {In}}:  {\itshape Proceedings of the IEEE
  Conference on Computer Vision and Pattern Recognition},  6154--6162.

\bibitem[\protect\citeauthoryear{Cao and Qiu}{2018}]{cao18}
Cao, R. and Qiu, G., 2018. Urban land use classification based on aerial and
  ground images. {\itshape {In}}:  {\itshape {International Conference on
  Content-Based Multimedia Indexing}},  1--6.

\bibitem[\protect\citeauthoryear{Cao {\itshape{et~al.}}}{2018}]{cao18a}
Cao, R., {\itshape et~al.}, 2018. Integrating aerial and street view images for
  urban land use classification. {\itshape {Remote Sensing}}, 10 (10), 1553.

\bibitem[\protect\citeauthoryear{Chang
  {\itshape{et~al.}}}{2020}]{chang2020mapping}
Chang, S., {\itshape et~al.}, 2020. Mapping the essential urban land use in
  Changchun by applying random forest and multi-source geospatial data.
  {\itshape Remote Sensing}, 12 (15), 2488.

\bibitem[\protect\citeauthoryear{Chen
  {\itshape{et~al.}}}{2018}]{chen2018encoder}
Chen, L.C., {\itshape et~al.}, 2018. Encoder-decoder with atrous separable
  convolution for semantic image segmentation. {\itshape {In}}:  {\itshape
  Proceedings of the European Conference on computer Vision},  801--818.

\bibitem[\protect\citeauthoryear{{Copernicus Land Monitoring
  Service}}{2020}]{Urban}
{Copernicus Land Monitoring Service}, 2020. {Urban Atlas}.  {Available from
  https://land.copernicus.eu/local/urban-atlas. [Accessed Nov. 2020]}.

\bibitem[\protect\citeauthoryear{Cordts {\itshape{et~al.}}}{2016}]{cordt16}
Cordts, M., {\itshape et~al.}, 2016. The cityscapes dataset for semantic urban
  scene understanding. {\itshape {In}}:  {\itshape {Proceedings of the IEEE
  Conference on Computer Vision and Pattern Recognition}},  3213--3223.

\bibitem[\protect\citeauthoryear{Defazio {\itshape{et~al.}}}{2014}]{defaz14}
Defazio, A., Bach, F.R., and Lacoste{-}Julien, S., 2014. {SAGA:} {A} fast
  incremental gradient method with support for non-strongly convex composite
  Objectives. {\itshape {CoRR}}, abs/1407.0202.

\bibitem[\protect\citeauthoryear{Deng {\itshape{et~al.}}}{2009}]{deng09}
Deng, J., {\itshape et~al.}, 2009. {Imagenet: A large-scale hierarchical image
  database}. {\itshape {In}}:  {\itshape {IEEE Conference on Computer Vision
  and Pattern Recognition}},  248--255.

\bibitem[\protect\citeauthoryear{Dustin}{2020}]{web7}
Dustin, S., 2020. Social media 2020: Top networks by the numbers [online].
  {Available from https://dustinstout.com/social-media-statistics/ [Accessed
  November 2020]}.

\bibitem[\protect\citeauthoryear{Estima and Painho}{2013}]{estim13}
Estima, J. and Painho, M., 2013. Exploratory analysis of {OpenStreetMap} for
  land use classification. {\itshape {In}}:  {\itshape {Proceedings of the
  Second ACM SIGSPATIAL International Workshop on Crowdsourced and Volunteered
  Geographic Information}},  39--46.

\bibitem[\protect\citeauthoryear{Estima and Painho}{2015}]{estim15}
Estima, J. and Painho, M., 2015. 13. {\itshape {In}}: {\itshape Investigating
  the potential of {OpenStreetMap} for land use/land cover production: A case
  study for continental {Portugal}}.,  273--293  Springer, Cham.

\bibitem[\protect\citeauthoryear{Fan {\itshape{et~al.}}}{2014}]{fan14}
Fan, H., Zipf, A., and Fu, Q., 2014. 2. {\itshape {In}}: {\itshape Estimation
  of building types on {OpenStreetMap} based on urban morphology analysis}.,
  19--35  Springer, Cham.

\bibitem[\protect\citeauthoryear{Fang {\itshape{et~al.}}}{2018}]{fang18}
Fang, F., {\itshape et~al.}, 2018. Urban land-use classification from
  photographs. {\itshape {{IEEE} Geoscience and Remote Sensing Letters}}, 15
  (12), 1927--1931.

\bibitem[\protect\citeauthoryear{Feng {\itshape{et~al.}}}{2018}]{feng18}
Feng, T., {\itshape et~al.}, 2018. Urban zoning using higher-order markov
  random fields on multi-view imagery data. {\itshape {In}}:  {\itshape
  {Proceedings of the European Conference on Computer Vision}},  614--630.

\bibitem[\protect\citeauthoryear{Fonte and Martinho}{2017}]{Font17}
Fonte, C.C. and Martinho, N., 2017. Assessing the applicability of
  OpenStreetMap data to assist the validation of land use/land cover maps.
  {\itshape International Journal of Geographical Information Science}, 31
  (12), 2382--2400.

\bibitem[\protect\citeauthoryear{Fuller and Moore}{2017}]{fulle17}
Fuller, M. and Moore, R., 2017. {\itshape {The death and life of great American
  cities}}.   Macat Library.

\bibitem[\protect\citeauthoryear{Gao {\itshape{et~al.}}}{2017}]{gao17}
Gao, S., Janowicz, K., and Couclelis, H., 2017. Extracting urban functional
  regions from points of interest and human activities on location-based social
  networks. {\itshape {Transactions in GIS}}, 21 (3), 446--467.

\bibitem[\protect\citeauthoryear{Haklay and Weber}{2008}]{hakla08}
Haklay, M. and Weber, P., 2008. {OpenStreetMap}: User-generated street maps.
  {\itshape {{IEEE} Pervasive Computing}}, 7 (4), 12--18.

\bibitem[\protect\citeauthoryear{He {\itshape{et~al.}}}{2016}]{he2016deep}
He, K., {\itshape et~al.}, 2016. Deep residual learning for image recognition.
  {\itshape {In}}:  {\itshape Proceedings of the IEEE Conference on Computer
  Vision and Pattern Recognition},  770--778.

\bibitem[\protect\citeauthoryear{{Hoffmann}
  {\itshape{et~al.}}}{2019}]{hoffm19b}
{Hoffmann}, E.J., {Werner}, M., and {Zhu}, X.X., 2019. Building instance
  classification using social media images. {\itshape {In}}:  {\itshape {Joint
  Urban Remote Sensing Event}},  1--4.

\bibitem[\protect\citeauthoryear{Hoffmann
  {\itshape{et~al.}}}{2019{\natexlab{a}}}]{hoffm19a}
Hoffmann, E.J., {\itshape et~al.}, 2019{\natexlab{a}}. Model fusion for
  building type classification from aerial and street view images. {\itshape
  {Remote Sensing}}, 11 (11), 1259.

\bibitem[\protect\citeauthoryear{Hoffmann
  {\itshape{et~al.}}}{2019{\natexlab{b}}}]{hoffm19}
Hoffmann, E.J., Werner, M., and Zhu, X., 2019{\natexlab{b}}. Mutual information
  analysis of social media images and building functions. {\itshape {In}}:
  {\itshape {IEEE International Geoscience and Remote Sensing Symposium}}.

\bibitem[\protect\citeauthoryear{Huang {\itshape{et~al.}}}{2020}]{huang20}
Huang, Z., {\itshape et~al.}, 2020. An ensemble learning approach for urban
  land use mapping based on remote sensing imagery and social sensing data.
  {\itshape Remote Sensing}, 12 (19), 3254.

\bibitem[\protect\citeauthoryear{Iovan {\itshape{et~al.}}}{2012}]{iovan12}
Iovan, C., {\itshape et~al.}, 2012. Classification of urban scenes from
  geo-referenced images in urban street-view context. {\itshape {In}}:
  {\itshape {11th International Conference on Machine Learning and
  Applications}}, Vol. ~2,  339--344.

\bibitem[\protect\citeauthoryear{Jiang {\itshape{et~al.}}}{2015}]{jiang15}
Jiang, S., {\itshape et~al.}, 2015. Mining point-of-interest data from social
  networks for urban land use classification and disaggregation. {\itshape
  {Computers, Environment and Urban Systems}}, 53, 36--46.

\bibitem[\protect\citeauthoryear{Jokar~Arsanjani
  {\itshape{et~al.}}}{2013}]{jokar13}
Jokar~Arsanjani, J., {\itshape et~al.}, 2013. Toward mapping land-use patterns
  from volunteered geographic information. {\itshape {International Journal of
  Geographical Information Science}}, 27 (12), 2264--2278.

\bibitem[\protect\citeauthoryear{Kang {\itshape{et~al.}}}{2018}]{kang18}
Kang, J., {\itshape et~al.}, 2018. Building instance classification using
  street view images. {\itshape {ISPRS Journal of Photogrammetry and Remote
  Sensing}}, 145, 44--59.

\bibitem[\protect\citeauthoryear{Karasov
  {\itshape{et~al.}}}{2019}]{karasov2019mapping}
Karasov, O., {\itshape et~al.}, 2019. Mapping the extent of land cover colour
  harmony based on satellite Earth observation data. {\itshape GeoJournal}, 84
  (4), 1057--1072.

\bibitem[\protect\citeauthoryear{Krizhevsky {\itshape{et~al.}}}{2012}]{krizh12}
Krizhevsky, A., Sutskever, I., and Hinton, G.E., 2012. Imagenet classification
  with deep convolutional neural networks. {\itshape {In}}:  {\itshape
  {Advances in Neural Information Processing Systems}},  1097--1105.

\bibitem[\protect\citeauthoryear{Lef{\`e}vre
  {\itshape{et~al.}}}{2017}]{lefev17}
Lef{\`e}vre, S., {\itshape et~al.}, 2017. Toward seamless multiview scene
  analysis from satellite to street level. {\itshape {Proceedings of the
  IEEE}}, 105 (10), 1884--1899.

\bibitem[\protect\citeauthoryear{Leung and Newsam}{2009}]{leung09}
Leung, D. and Newsam, S., 2009. Proximate sensing using georeferenced community
  contributed photo collections. {\itshape {In}}:  {\itshape {Proceedings of
  the 2009 International Workshop on Location Based Social Networks}},  57--64.

\bibitem[\protect\citeauthoryear{Leung and Newsam}{2012}]{leung12}
Leung, D. and Newsam, S., 2012. Exploring geotagged images for land-use
  classification. {\itshape {In}}:  {\itshape {Proceedings of the ACM
  Multimedia 2012 Workshop on Geotagging and Its Applications in Multimedia}},
  3--8.

\bibitem[\protect\citeauthoryear{Li {\itshape{et~al.}}}{2010}]{li10}
Li, L.J., {\itshape et~al.}, 2010. Object bank: A high-level image
  representation for scene classification \& semantic feature sparsification.
  {\itshape {In}}:  {\itshape {Advances in Neural Information Processing
  Systems}},  1378--1386.

\bibitem[\protect\citeauthoryear{Li and Zhang}{2016}]{li16}
Li, X. and Zhang, C., 2016. Urban land use information retrieval based on scene
  classification of {Google Street View} images.. {\itshape {In}}:  {\itshape
  {SDW@ GIScience}},  41--46.

\bibitem[\protect\citeauthoryear{Lin {\itshape{et~al.}}}{2014}]{lin14}
Lin, T.Y., {\itshape et~al.}, 2014. {Microsoft COCO: Common objects in
  context}. {\itshape {In}}:  {\itshape {European Conference on Computer
  Vision}},  740--755.

\bibitem[\protect\citeauthoryear{Liu {\itshape{et~al.}}}{2020}]{liu2020annual}
Liu, D., {\itshape et~al.}, 2020. Annual large-scale urban land mapping based
  on Landsat time series in Google Earth Engine and OpenStreetMap data: A case
  study in the middle Yangtze River basin. {\itshape ISPRS Journal of
  Photogrammetry and Remote Sensing}, 159, 337--351.

\bibitem[\protect\citeauthoryear{Liu {\itshape{et~al.}}}{2016}]{liu16}
Liu, W., {\itshape et~al.}, 2016. {SSD: Single shot multibox detector}.
  {\itshape {In}}:  {\itshape {European Conference on Computer Vision}},
  21--37.

\bibitem[\protect\citeauthoryear{Machado
  {\itshape{et~al.}}}{2020}]{machado2020airound}
Machado, G., {\itshape et~al.}, 2020. AiRound and CV-BrCT: Novel multi-view
  datasets for scene classification. {\itshape arXiv preprint
  arXiv:2008.01133}.

\bibitem[\protect\citeauthoryear{Mahabir
  {\itshape{et~al.}}}{2020{\natexlab{a}}}]{mahabir2020crowdsourcing}
Mahabir, R., {\itshape et~al.}, 2020{\natexlab{a}}. Crowdsourcing street view
  imagery: A comparison of Mapillary and OpenStreetCam. {\itshape ISPRS
  International Journal of Geo-Information}, 9 (6), 341.

\bibitem[\protect\citeauthoryear{Mahabir
  {\itshape{et~al.}}}{2020{\natexlab{b}}}]{Maha20}
Mahabir, R., {\itshape et~al.}, 2020{\natexlab{b}}. Crowdsourcing Street View
  Imagery: A Comparison of Mapillary and OpenStreetCam. {\itshape ISPRS
  International Journal of Geo-Information}, 9 (6).

\bibitem[\protect\citeauthoryear{{Ministry of Infrastructure and the
  Environment}}{2020}]{web1}
{Ministry of Infrastructure and the Environment}, 2020. {Addresses and
  buildings key register [online]}.  {Available from
  https://business.gov.nl/regulation/ addresses-and-buildings-key-geo-register
  [Accessed November 2020]}.

\bibitem[\protect\citeauthoryear{Movshovitz-Attias
  {\itshape{et~al.}}}{2015}]{movsh15}
Movshovitz-Attias, Y., {\itshape et~al.}, 2015. Ontological supervision for
  fine grained classification of street view storefronts. {\itshape {In}}:
  {\itshape {Proceedings of the IEEE Conference on Computer Vision and Pattern
  Recognition}},  1693--1702.

\bibitem[\protect\citeauthoryear{Munoz
  {\itshape{et~al.}}}{2020}]{munoz2020openstreetmap}
Munoz, J.E.V., {\itshape et~al.}, 2020. OpenStreetMap: Challenges and
  opportunities in machine learning and remote sensing. {\itshape IEEE
  Geoscience and Remote Sensing Magazine}.

\bibitem[\protect\citeauthoryear{Neuhold {\itshape{et~al.}}}{2017}]{neuho17}
Neuhold, G., {\itshape et~al.}, 2017. The Mapillary Vistas dataset for semantic
  understanding of street scenes. {\itshape {In}}:  {\itshape {Proceedings of
  the IEEE International Conference on Computer Vision}},  4990--4999.

\bibitem[\protect\citeauthoryear{Penatti {\itshape{et~al.}}}{2015}]{penat15}
Penatti, O.A., Nogueira, K., and Dos~Santos, J.A., 2015. Do deep features
  generalize from everyday objects to remote sensing and aerial scenes
  domains?. {\itshape {In}}:  {\itshape {Proceedings of the IEEE Conference on
  Computer Vision and Pattern Recognition Workshops}},  44--51.

\bibitem[\protect\citeauthoryear{Qiao {\itshape{et~al.}}}{2020}]{Qiao20a}
Qiao, Z., Yuan, X., and Elhoseny, M., 2020. Urban scene recognition via deep
  network integration. {\itshape {In}}:  {\itshape International Conference on
  Urban Intelligence and Applications},  135--149.

\bibitem[\protect\citeauthoryear{Qiao {\itshape{et~al.}}}{2021}]{Qiao20b}
Qiao, Z., {\itshape et~al.}, 2021. Attention pyramid module for scene
  Recognition. {\itshape {In}}:  {\itshape International Conference on Pattern
  Recognition 2020}, Jan. 10-15., Milan, Italy.

\bibitem[\protect\citeauthoryear{Redmon and Farhadi}{2018}]{redmon2018yolov3}
Redmon, J. and Farhadi, A., 2018. Yolov3: An incremental improvement. {\itshape
  arXiv preprint arXiv:1804.02767}.

\bibitem[\protect\citeauthoryear{Ren {\itshape{et~al.}}}{2015}]{ren15}
Ren, S., {\itshape et~al.}, 2015. Faster r-cnn: Towards real-time object
  detection with region proposal networks. {\itshape {In}}:  {\itshape
  {Advances in Neural Information Pprocessing Systems}},  91--99.

\bibitem[\protect\citeauthoryear{Rupali and Patil}{2016}]{rupal16}
Rupali, A.S. and Patil, D., 2016. A mechanism for learning and recognition of
  on-premise signs from street view images. {\itshape {In}}:  {\itshape
  {Symposium on Colossal Data Analysis and Networking}},  1--4.

\bibitem[\protect\citeauthoryear{S{\"a}yn{\"a}joki
  {\itshape{et~al.}}}{2014}]{saynaj14}
S{\"a}yn{\"a}joki, E.S., Heinonen, J., and Junnila, S., 2014. The power of
  urban planning on environmental sustainability: A focus group study in
  Finland. {\itshape {Sustainability}}, 6 (10), 6622--6643.

\bibitem[\protect\citeauthoryear{Sharifi~Noorian
  {\itshape{et~al.}}}{2020}]{sharifi2020detecting}
Sharifi~Noorian, S., {\itshape et~al.}, 2020. Detecting, classifying, and
  mapping retail storefronts using street-level imagery. {\itshape {In}}:
  {\itshape Proceedings of the 2020 International Conference on Multimedia
  Retrieval},  495--501.

\bibitem[\protect\citeauthoryear{Shrivastava
  {\itshape{et~al.}}}{2016}]{shriv16}
Shrivastava, A., Gupta, A., and Girshick, R., 2016. Training region-based
  object detectors with online hard example mining. {\itshape {In}}:  {\itshape
  {Proceedings of the IEEE Conference on Computer Vision and Pattern
  Recognition}},  761--769.

\bibitem[\protect\citeauthoryear{Simonyan and Zisserman}{2014}]{simon14}
Simonyan, K. and Zisserman, A., 2014. Very deep convolutional networks for
  large-scale image recognition. {\itshape arXiv preprint arXiv:1409.1556}.

\bibitem[\protect\citeauthoryear{Srivastava
  {\itshape{et~al.}}}{2018{\natexlab{a}}}]{srivas18a}
Srivastava, S., {\itshape et~al.}, 2018{\natexlab{a}}. Land-use
  characterisation using {Google Street View} pictures and {OpenStreetMap}.
  {\itshape {In}}:  {\itshape {Proceedings of the Association of Geographic
  Information Laboratories in Europe Conference}}.

\bibitem[\protect\citeauthoryear{Srivastava
  {\itshape{et~al.}}}{2020}]{sriva18c}
Srivastava, S., {\itshape et~al.}, 2020. Fine-grained landuse characterization
  using ground-based pictures: {A} deep learning solution based on globally
  available data. {\itshape {International Journal of Geographical Information
  Science}},  1--20.

\bibitem[\protect\citeauthoryear{Srivastava
  {\itshape{et~al.}}}{2018{\natexlab{b}}}]{srivas18}
Srivastava, S., {\itshape et~al.}, 2018{\natexlab{b}}. Multilabel building
  functions classification from ground pictures using convolutional neural
  networks. {\itshape {In}}:  {\itshape {Proceedings of the 2nd ACM SIGSPATIAL
  International Workshop on AI for Geographic Knowledge Discovery}},  43--46.

\bibitem[\protect\citeauthoryear{Srivastava
  {\itshape{et~al.}}}{2019}]{sriva19c}
Srivastava, S., Vargas-Mu{\~n}oz, J.E., and Tuia, D., 2019. Understanding urban
  landuse from the above and ground perspectives: A deep learning, multimodal
  solution. {\itshape {Remote Sensing of Environment}}, 228, 129--143.

\bibitem[\protect\citeauthoryear{Terroso-Saenz
  {\itshape{et~al.}}}{2021}]{Terr21}
Terroso-Saenz, F., Mu\~{n}oz, A., and Arcas, F., 2021. Land-use dynamic
  discovery based on heterogeneous mobility sources. {\itshape International
  Journal of Intelligent Systems}, 36 (1), 478--525.

\bibitem[\protect\citeauthoryear{Tian {\itshape{et~al.}}}{2017}]{tian17}
Tian, Y., Chen, C., and Shah, M., 2017. Cross-view image matching for
  geo-localization in urban environments. {\itshape {In}}:  {\itshape
  {Proceedings of the {IEEE} Conference on Computer Vision and Pattern
  Recognition}},  3608--3616.

\bibitem[\protect\citeauthoryear{Tracewski {\itshape{et~al.}}}{2017}]{trace17}
Tracewski, L., Bastin, L., and Fonte, C.C., 2017. Repurposing a deep learning
  network to filter and classify volunteered photographs for land cover and
  land use characterization. {\itshape {Geo-spatial Information Science}}, 20
  (3), 252--268.

\bibitem[\protect\citeauthoryear{Tsai {\itshape{et~al.}}}{2014}]{tsai14}
Tsai, T.H., {\itshape et~al.}, 2014. Learning and recognition of on-premise
  signs from weakly labeled street view images. {\itshape {IEEE Transactions on
  Image Processing}}, 23 (3), 1047--1059.

\bibitem[\protect\citeauthoryear{Van De~Sande {\itshape{et~al.}}}{2009}]{van09}
Van De~Sande, K., Gevers, T., and Snoek, C., 2009. Evaluating color descriptors
  for object and scene recognition. {\itshape {IEEE Transactions on Pattern
  Analysis and Machine Intelligence}}, 32 (9), 1582--1596.

\bibitem[\protect\citeauthoryear{Vargas Mu\~{n}oz
  {\itshape{et~al.}}}{2020}]{Varg20}
Vargas Mu\~{n}oz, J.E., {\itshape et~al.}, 2020. OpenStreetMap: Challenges and
  Opportunities in Machine Learning and Remote Sensing. {\itshape IEEE
  Geoscience and Remote Sensing Magazine},  0--0.

\bibitem[\protect\citeauthoryear{Vargas-Mu{\~n}oz
  {\itshape{et~al.}}}{2019}]{varga19}
Vargas-Mu{\~n}oz, J.E., {\itshape et~al.}, 2019. Correcting rural building
  annotations in {OpenStreetMap} using convolutional neural networks. {\itshape
  {ISPRS Journal of Photogrammetry and Remote Sensing}}, 147, 283--293.

\bibitem[\protect\citeauthoryear{Wang {\itshape{et~al.}}}{2017}]{wang17a}
Wang, Q., Zhou, C., and Xu, N., 2017. Street view image classification based on
  convolutional neural network. {\itshape {In}}:  {\itshape {IEEE 2nd Advanced
  Information Technology, Electronic and Automation Control Conference}},
  1439--1443.

\bibitem[\protect\citeauthoryear{Wang and Hofe}{2008}]{wang08}
Wang, X. and Hofe, R., 2008. {\itshape Research methods in urban and regional
  planning}.   {Springer Science \& Business Media}.

\bibitem[\protect\citeauthoryear{{Wikipedia}}{2020}]{wiki2}
{Wikipedia}, 2020. {Coverage of {Google Street View [online]}}.  {Available
  from https://en.wikipedia.org/w/index.php?\&oldid=909938748 [Accessed
  November 2020]}.

\bibitem[\protect\citeauthoryear{Workman {\itshape{et~al.}}}{2017}]{workm17}
Workman, S., {\itshape et~al.}, 2017. A unified model for near and remote
  sensing. {\itshape {In}}:  {\itshape {Proceedings of the {IEEE} International
  Conference on Computer Vision}},  2688--2697.

\bibitem[\protect\citeauthoryear{Xiao {\itshape{et~al.}}}{2010}]{xiao10}
Xiao, J., {\itshape et~al.}, 2010. Sun database: Large-scale scene recognition
  from abbey to zoo. {\itshape {In}}:  {\itshape {{IEEE} Computer Society
  Conference on Computer Vision and Pattern Recognition}},  3485--3492.

\bibitem[\protect\citeauthoryear{Ye {\itshape{et~al.}}}{2019}]{ye19}
Ye, Y., {\itshape et~al.}, 2019. Land use classification from social media data
  and satellite imagery. {\itshape {The Journal of Supercomputing}},  1--16.

\bibitem[\protect\citeauthoryear{You {\itshape{et~al.}}}{2015}]{you15}
You, Q., {\itshape et~al.}, 2015. Robust image sentiment analysis using
  progressively trained and domain transferred deep networks. {\itshape {In}}:
  {\itshape {Twenty-ninth AAAI Conference on Artificial Intelligence}}.

\bibitem[\protect\citeauthoryear{Yuan and Sarma}{2011}]{Yuan11}
Yuan, X. and Sarma, V., 2011. Automatic urban water-body detection and
  segmentation from sparse ALSM data via spatially constrained model-driven
  clustering. {\itshape {IEEE Geoscience and Remote Sensing Letters}}, 8 (1),
  73--77.

\bibitem[\protect\citeauthoryear{Yuan {\itshape{et~al.}}}{2002}]{Yuan02}
Yuan, X., {\itshape et~al.}, 2002. Mining negative association rules. {\itshape
  {In}}:  {\itshape {Proceedings ISCC 2002 Seventh International Symposium on
  Computers and Communications}}, July.,  623--628.

\bibitem[\protect\citeauthoryear{Zamir {\itshape{et~al.}}}{2011}]{zamir11}
Zamir, A.R., Darino, A., and Shah, M., 2011. Street view challenge:
  Identification of commercial entities in street view imagery. {\itshape
  {In}}:  {\itshape {10th International Conference on Machine Learning and
  Applications and Workshops}}, Vol. ~2,  380--383.

\bibitem[\protect\citeauthoryear{Zhang
  {\itshape{et~al.}}}{2017{\natexlab{a}}}]{zhang17a}
Zhang, W., {\itshape et~al.}, 2017{\natexlab{a}}. Parcel-based urban land use
  classification in megacity using airborne LiDAR, high resolution
  orthoimagery, and {Google Street View}. {\itshape {Computers, Environment and
  Urban Systems}}, 64, 215--228.

\bibitem[\protect\citeauthoryear{Zhang
  {\itshape{et~al.}}}{2017{\natexlab{b}}}]{zhang17}
Zhang, W., {\itshape et~al.}, 2017{\natexlab{b}}. Parcel feature data derived
  from {Google Street View} images for urban land use classification in
  {Brooklyn, New York City for urban land use classification in Brooklyn, New
  York City}. {\itshape {Data in Brief}}, 12, 175--179.

\bibitem[\protect\citeauthoryear{Zhang {\itshape{et~al.}}}{2010}]{zhang10}
Zhang, Y., Jin, R., and Zhou, Z.H., 2010. Understanding bag-of-words model: A
  statistical framework. {\itshape {International Journal of Machine Learning
  and Cybernetics}}, 1 (1-4), 43--52.

\bibitem[\protect\citeauthoryear{Zhao
  {\itshape{et~al.}}}{2020}]{zhao2020bounding}
Zhao, K., {\itshape et~al.}, 2020. Bounding boxes are all we need: Street view
  image classification via context encoding of detected buildings. {\itshape
  arXiv preprint arXiv:2010.01305}.

\bibitem[\protect\citeauthoryear{Zhou {\itshape{et~al.}}}{2016}]{zhou16}
Zhou, B., {\itshape et~al.}, 2016. Places: An image database for deep scene
  understanding. {\itshape arXiv preprint arXiv:1610.02055}.

\bibitem[\protect\citeauthoryear{Zhou {\itshape{et~al.}}}{2017}]{zhou17}
Zhou, B., {\itshape et~al.}, 2017. Places: A 10 million image database for
  scene recognition. {\itshape {{IEEE} Transactions on Pattern Analysis and
  Machine Intelligence}}, 40 (6), 1452--1464.

\bibitem[\protect\citeauthoryear{Zhou {\itshape{et~al.}}}{2014}]{zhou14}
Zhou, B., {\itshape et~al.}, 2014. Learning deep features for scene recognition
  using places database. {\itshape {In}}:  {\itshape {Advances in Neural
  Information Processing Systems}},  487--495.

\bibitem[\protect\citeauthoryear{Zhu {\itshape{et~al.}}}{2019}]{zhu19}
Zhu, Y., Deng, X., and Newsam, S., 2019. Fine-grained land use classification
  at the city scale using ground-level images. {\itshape {{IEEE} Transactions
  on Multimedia}}.

\bibitem[\protect\citeauthoryear{Zhu and Newsam}{2015}]{zhu15}
Zhu, Y. and Newsam, S., 2015. Land use classification using convolutional
  neural networks applied to ground-level images. {\itshape {In}}:  {\itshape
  {Proceedings of the 23rd SIGSPATIAL International Conference on Advances in
  Geographic Information Systems}}, p.~61.

\end{thebibliography}

\label{lastpage}

\end{document}